\documentclass{article}

    \usepackage[nonatbib, preprint]{neurips_2019}

\usepackage[utf8]{inputenc} 
\usepackage[T1]{fontenc}    
\usepackage{hyperref}       
\usepackage{url}            
\usepackage{booktabs}       
\usepackage{amsfonts}       
\usepackage{nicefrac}       
\usepackage{microtype}      

\usepackage{graphicx}
\usepackage{subcaption}
\usepackage{mwe}
\usepackage{amsmath}
\usepackage{amsthm}
\usepackage{common}

\theoremstyle{plain}
\newtheorem{theorem}{Theorem}
\newtheorem{lemma}{Lemma}
\newtheorem{definition}{Definition}

\title{CheckNet: Secure Inference on Untrusted Devices}

\author{%
  Marcus Comiter\\
  Harvard University\\
  Cambridge, MA \\
  \texttt{marcuscomiter@g.harvard.edu} \\
   \And
   Surat Teerapittayanon \\
   Harvard University \\
   Cambridge, MA \\
   \texttt{steerapi@seas.harvard.edu} \\
   \AND
   H.T. Kung \\
   Harvard University \\
   Cambridge, MA \\
   \texttt{kung@harvard.edu} \\
}

\begin{document}

\maketitle

\begin{abstract}
We introduce CheckNet, a method for secure inference with deep neural networks on untrusted devices.  CheckNet is like a checksum for neural network inference: it verifies the integrity of the inference computation performed by untrusted devices to 1) ensure the inference has actually been performed, and 2) ensure the inference has not been manipulated by an attacker. CheckNet is completely transparent to the third party running the computation, applicable to all types of neural networks, does not require specialized hardware, adds little overhead, and has negligible impact on model performance.  CheckNet can be configured to provide different levels of security depending on application needs and compute/communication budgets.  We present both empirical and theoretical validation of CheckNet on multiple popular deep neural network models, showing excellent attack detection (0.88-0.99 AUC) and attack success bounds.
\end{abstract}

\section{Introduction}

Leveraging untrusted third party devices and resources to execute deep learning inference serves an important role, but poses serious security risks.  Untrusted third parties can purposely manipulate inference outputs in order to attack the machine learning application, for example, by changing the result of a classification problem. Alternatively, third parties can be ``lazy'' and return random or previously computed outputs in order to avoid expending computational resources performing the actual inference calculation. 

In order to allow for \textit{trusted} inference computation on \textit{untrusted} third party devices, there is a need to verify the integrity of the inference computation.  Limited efforts to date in this field have restricted applicability or required specialized hardware.  New methods are needed that are generally applicable to commonly used and deployed deep learning models.

To this end we introduce CheckNet, a general method for verifying the integrity of deep learning inference computations performed by an untrusted third party. Just as a checksum verifies the integrity of data, CheckNet verifies the integrity of a machine learning inference computation.  Specifically, it verifies that a given inference computation has been correctly executed and that the results have not been manipulated.  CheckNet allows for configurable security levels based on application needs while being applicable to a wide range of deep neural network models.

In order to be widely applicable and easily used, CheckNet is designed to be compatible with all common deep learning methods and not require any specialized hardware.  Specifically, CheckNet has the following design goals: 1) fast to verify the integrity of the computation and output; 2) requires only minimal change and retraining (last layer only) of the original network; and 3) adds only minimal communication and computation overhead. 

CheckNet verifies inference integrity with two techniques.  The first, HashCheck, verifies that the inference computation is consistent with respect to the input.  The second, CrossCheck, verifies that the inference output has not been manipulated.  Together, these techniques are able to protect against attacks such as replay attacks (adversary returns an old but valid inference result), random attacks (adversary returns random results in order to avoid actually performing the inference computation), and targeted manipulation attacks (adversary changes the values of particular output nodes in order to change the resulting classification).  Further, CheckNet is completely transparent to the third party.  While the classification task is used in this paper, CheckNet is applicable to other deep learning tasks, as it acts on node values of the penultimate layer (e.g., prior to the softmax layer), therefore not tying it to a specific task.

We present empirical and theoretical validation of CheckNet on multiple popular deep neural network models under a wide set of attacks, showing excellent attack detection capabilities, and prove bounds on the probability of inference attacks evading CheckNet detection.

 



\section{CheckNet}
CheckNet is a general method to verify the integrity of an inference computation. Given an original network computation $\boldy = t(\boldx)$ for network $t$, CheckNet verifies for a given input $\boldx$ that the inference computation $t(\boldx)$ has actually been executed on $\boldx$, and that the output $\boldy$ has not been manipulated.  To do this, CheckNet uses two techniques: HashCheck and CrossCheck.  HashCheck ensures the inference computation has actually been executed on $\boldx$.  It does this by constructing a pair of hash functions $(f,g)$ such that the bithash $\boldh_{x} = f(\boldx)$ of inference input $\boldx$ is approximately equal to the bithash $\boldh_{y} = g(\boldy)$ of inference output $\boldy$, or $\boldh_{x} \approx \boldh_{y}$.  CrossCheck ensures the output has not been manipulated (e.g., by swapping output node values to change the classification result). It does this by obscuring the result of the model by outputting multiple intertwined sets of results, building in redundancy, and cross checking the results among themselves.  Figure~\ref{fig:CheckNetFig} shows an overview of CheckNet and its two techniques.

\subsection{HashCheck}

HashCheck ensures that the output returned from an inference calculation is consistent with respect to the input. This helps protects against attacks in which, for either adversarial or lazy reasons, third parties purposely return an output that is unrelated to the input. Examples include returning either an old but valid inference result, or randomly generated inference result, instead of the true inference result.  An overview of HashCheck is shown at the top of Figure~\ref{fig:CheckNetFig}.

HashCheck provides a fast integrity check between the input and output by constructing a pair of hash functions $(f,g)$ relating the input $\boldx$ and the inference output $\boldy$ to a common bithash value. Specifically, hash functions are built such that the bithash $\boldh_{x} = f(\boldx)$ of inference input $\boldx$ is approximately equal (within a threshold) to the bithash $\boldh_{y} = g(\boldy)$ of inference output $\boldy$, or $\boldh_{x} \approx \boldh_{y}$.  Further, the hash functions are built to have negligible computation cost compared with the network inference itself. 
In this paper, we use an MLP with one hidden layer whose output is a binary code as the hash functions.  For the results in this paper with AlexNet~\cite{krizhevsky2012imagenet} on the 10-class classification problem, verification with HashCheck uses only approximately 1\% of the MACs used for inference and only $2.9$kB of additional communication.

The HashCheck training process involves no modifications to the underlying model and uses a dataset created by pairing the inputs $\boldX = \{\boldx\}_{i=1}^N$ with the output $\boldY = \{\boldy\}_{i=1}^N$ of the model, where $N$ is the size of the dataset.  The two hash networks are derived in the following process: first, the $g$ network is initialized as a random matrix followed by a binary quantization, therefore performing a random linear projection from $\boldy$.  This network remains fixed throughout the entire training process and serves as a source of randomness.  Second, the $f$ network is initialized as a single layer MLP followed by a sigmoid function, and is trained to map the input $\boldx$ to the bithash of the corresponding inference output $\boldy$.  $f$ is trained via backpropagation with loss term $L(\boldh_{x}, \boldh_{y})$, where $L$ is binary cross entropy loss, and $\boldh_{y}$ is the output of $g(\boldy)$ with binary quantization applied. Both hashes are of length $l$.  Through this process, $f$ learns to map the input $\boldx$ to the bithash code derived from the random projection of the output $\boldy$, as determined by $g$. Since random projections preserve distances~\cite{charikar2002similarity}, HashCheck learns the hash function $f$ mapping $\boldx$ into a bithash space where the distances between the outputs of the network $(\boldy = t(\boldx))$ are preserved. Any deviation from the true $\boldy$ will result in a mismatch between the bithash codes.  In order to increase robustness, multiple sets of $(f,g)$ functions are learned and used together.  Section~\ref{subsec:attackdetecmethod} describes how HashCheck detects attacks.

\subsection{CrossCheck}\label{subsec:crosscheck}

\begin{figure}[ht!]
    \centering
        \includegraphics[width=0.75\linewidth]{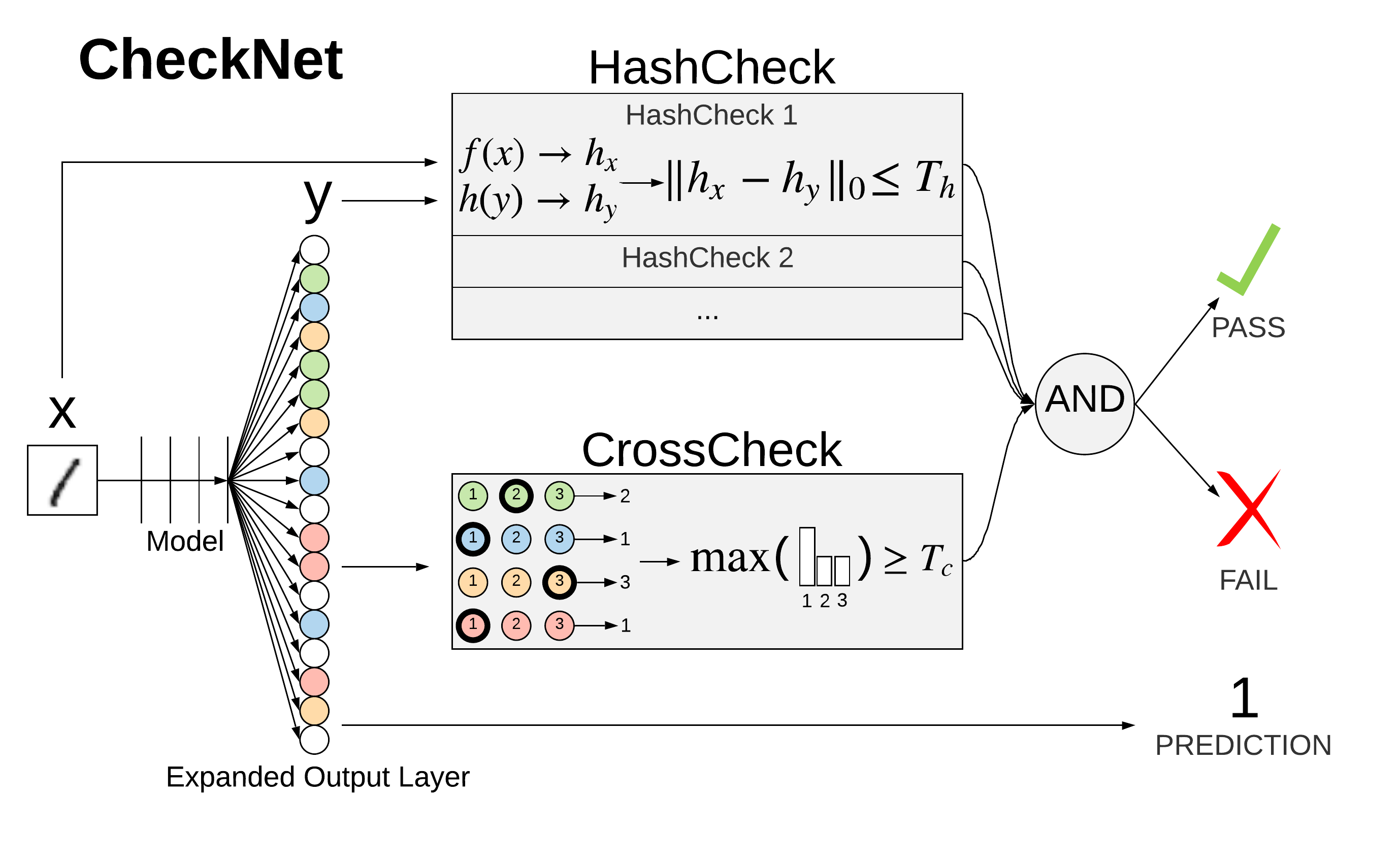}
        \caption{CheckNet verifies the integrity of inference computations using two methods: HashCheck (ensures result $\boldy$ is consistent with input $\boldx$)  and CrossCheck (ensures result has not been manipulated).}
        \label{fig:CheckNetFig}
\end{figure}

CrossCheck ensures the true output of an inference calculation has not been manipulated or fabricated in any way that changes the prediction outcome. It does this by obscuring the original output through returning multiple redundant, shuffled sets of results, and then cross checking the results among themselves, which delivers additional robustness as well.  An overview of CrossCheck is shown at the bottom of Figure~\ref{fig:CheckNetFig}.

To use CrossCheck, only the final layer of a model is modified and retrained to use CrossCheck. The process has three steps. First, the output layer is expanded to have additional output nodes. For a model performing $N_c$-class classification, the output layer is expanded to have $N_o$ output nodes, where $N_o\gg N_c$ (shown as the ``Expanded Output Layer'' in Figure~\ref{fig:CheckNetFig}, where the output layer of the model for a 3-class classification problem is expanded to have $20 \gg 3$ nodes). Second, $N_s$ CrossCheck sets are formed, where each CrossCheck set (shown as the green, blue, orange and red sets of nodes in Figure~\ref{fig:CheckNetFig}) consists of $N_c$ nodes chosen at random from the $N_o$ output nodes in the augmented model. CrossCheck sets may overlap (i.e., the same output node may be used in different CrossCheck sets, and may correspond to different labels in the different CrossCheck sets), and some nodes do not need to be in any CrossCheck sets at all, serving as decoys.  Only the model owner knows the CrossCheck sets.  The third party has no knowledge of how many sets are used or membership of these sets.  Third, only the last layer of the model is retrained on the training set such that each CrossCheck set correctly performs the $N_c$-class classification task using its own nodes. 
As a result of training the modified model in this way, each CrossCheck set should independently output the correct classification, and therefore can be used to ``cross check'' each other, as discussed in Section~\ref{subsec:attackdetecmethod}.  Changing the model in this way has no/negligible impact on accuracy.


This CrossCheck mechanism provides security by making it difficult for an adversary to manipulate the output from the model. For example, in a model without CrossCheck, if an adversary wished to change the model output, it could swap the maximum value in the output with another node, resulting in a different classification result. Under CrossCheck however, this attack would be difficult to execute because the adversary does not know which nodes to attack (because of obfuscation), how many nodes need to be attacked (because the level of redundancy is unknown), and the fact that mistakes are likely to be detected.  Even in cases where the adversary learned information about the sets, efforts to exploit this information will still be detected by CheckNet, as shown in Section~\ref{sec:results}.



\subsection{CheckNet Attack Detection Mechanism}\label{subsec:attackdetecmethod}
Figure~\ref{fig:CheckNetFig} shows how CheckNet uses the HashCheck and CrossCheck techniques.  Specifically, to use CheckNet to protect a model against attacks, the original model is modified to include the CrossCheck output layer, and a set of $N_h$ HashCheck hash functions $\{(f_i,g_i)\}_{i=1}^{N_h}$ are learned.  The CheckNet-protected model is then given to the untrusted third party, which uses the model in normal fashion.  The use of CheckNet is completely transparent to the third party.  
As far as the third party is concerned, the model is a normal model with an $N_o$-dimensional output. Specifically, the third party does not know the hash functions, the true underlying number of classes $N_c$ in the original model, the number of CrossCheck sets $N_s$, or the membership of each of the $N_s$ CrossCheck sets. 
The third party runs the inference normally, resulting in a $N_o$-dimensional output.

This $N_o$-dimensional output is then returned.  First, an ``unverified result'' is obtained by collecting the classification vote of each CrossCheck set (i.e., applying a softmax to each CrossCheck set and selecting the class corresponding to the maximum) and using a majority vote.  Second, the integrity of the result is verified using both HashCheck and CrossCheck.  For the HashCheck verification, for each pair of hash functions $(f_i,g_i)$, the bithash of the original input $\boldh_{x} = f_i(\boldx)$ and the bithash of the returned output $\boldh_{y}=g_i(\boldy)$ are obtained, and the bitwise distance between the two bithash values $\|\boldh_x - \boldh_y\|_0$ is compared to a threshold $T_h$.  If $\|\boldh_x - \boldh_y\|_0\leq T_h$ for some threshold $T_h$, the result passes that particular HashCheck integrity check.  For the CrossCheck verification, cardinality of the majority vote $m$ is compared with a threshold $T_c$.  If $m \geq T_c$, the result passes the CrossCheck integrity check.  If the result passes all of the HashCheck and CrossCheck integrity checks, the result is accepted. Otherwise, it is rejected. 

Both techniques are needed to detect the wide range of possible attacks, which include ``hard working/adversarial'' attacks, which seek to subvert the system by changing the prediction outcome, and ``lazy'' attacks, which seek to avoid actually executing the inference computation (see Section~\ref{sec:attacks} for examples of these attacks).  HashCheck alone cannot detect small changes in the output nodes, as certain margins of changes to the output are needed to change the resulting bithash (as neural networks are robust to these small changes).  CrossCheck provides the mechanism to thwart a ``hard-working'' adversary who may otherwise try to find tiny surfaces on the manifold that they can alter in order to subvert the system.  Further, CrossCheck alone does not protect against attacks that return valid outputs that do not match the input, as its job is to detect if an output is valid or not, and so relies on HashCheck to detect the discrepancy with the input.

The computational and communication complexity of CheckNet are both small compared to that of the original model.  The computational complexity of HashCheck is $O(N_hlN_o)$, where $N_h$ is the number of HashCheck networks, l is bithash length, and $N_o$ is the number of output nodes.  The computational complexity of the protected model's new last layer after being expanded by CrossCheck is $O(N_op)$, where $p$ is the number of output nodes in the model's penultimate layer, as compared to the $O(N_cp)$ complexity of the unprotected model's last layer.  Further, the communication complexity of CheckNet communication is $O(N_o)$, where $N_o$ is the number of output nodes, while the complexity of the unprotected model is $O(N_c)$.  These hyperparameters can be adjusted to meet compute and communication budgets, as is further discussed in Section~\ref{sec:results}.

\section{Evaluations\label{sec:results}}
We present empirical results for CheckNet under different attack models.  We evaluate on multiple popular models, MobileNet~\cite{howard2017mobilenets} and AlexNet~\cite{krizhevsky2012imagenet}, and evaluate performance using the CIFAR-10 dataset~\cite{krizhevsky2009learning}.  As MobileNet is a network specifically designed for mobile devices likely to run inference calculations for peers, these results are particularly salient. For all results, we use standard CIFAR-10 train/test splits (60k/10k samples, respectively).  We also get results on the FashionMNIST~\cite{xiao2017fashion} and CIFAR-100~\cite{krizhevsky2009learning} datasets, but observe similar trends and therefore omit them for space purposes.  We train the models and generate results using a GeForce RTX 2080 Ti GPU.


\subsection{Attacks Models\label{sec:attacks}}

We evaluate the effectiveness of CheckNet under three attack models on an inference computation $t(\boldx)$ computed on input $\boldx$ that should return $\boldy$:
\begin{enumerate}
    \item \textbf{Random:} a random inference result is returned, where the value of each output node is sampled from a distribution characterizing valid output node values.  This attack would be used by a ``lazy'' third party not wanting to actually run the inference computation.
    \item \textbf{Targeted Classification Change:} the value of $n$ output nodes within output vector $\boldy$ are changed to above the maximum value $\max(\boldy)$ in an attempt to change the classification result.  This attack would be used by an ``adversarial'' third party to change the result. Since this requires them to first run the original inference in order to obtain the result and then try to alter it, this attack requires effort, making it a ``hard-working'' third party attack.
    \item \textbf{Replay:} a different but otherwise valid inference result $\boldy'$ computed on a different $\boldx'$ is returned.  This attack would also be used by a ``lazy'' third party.  Additionally, if an ``adversarial'' third party was somehow able to learn about the number or membership of CrossCheck sets, their targeted classification attack would resemble a replay attack, as they would be able to closely approximate the output distribution of another class. 
    
    
\end{enumerate}

\subsection{Attack Detection and Receiver Operating Characteristic (ROC) Curves}
\begin{figure}[ht!]
    \centering
    \begin{minipage}[t]{0.47\textwidth}
        \includegraphics[width=\linewidth]{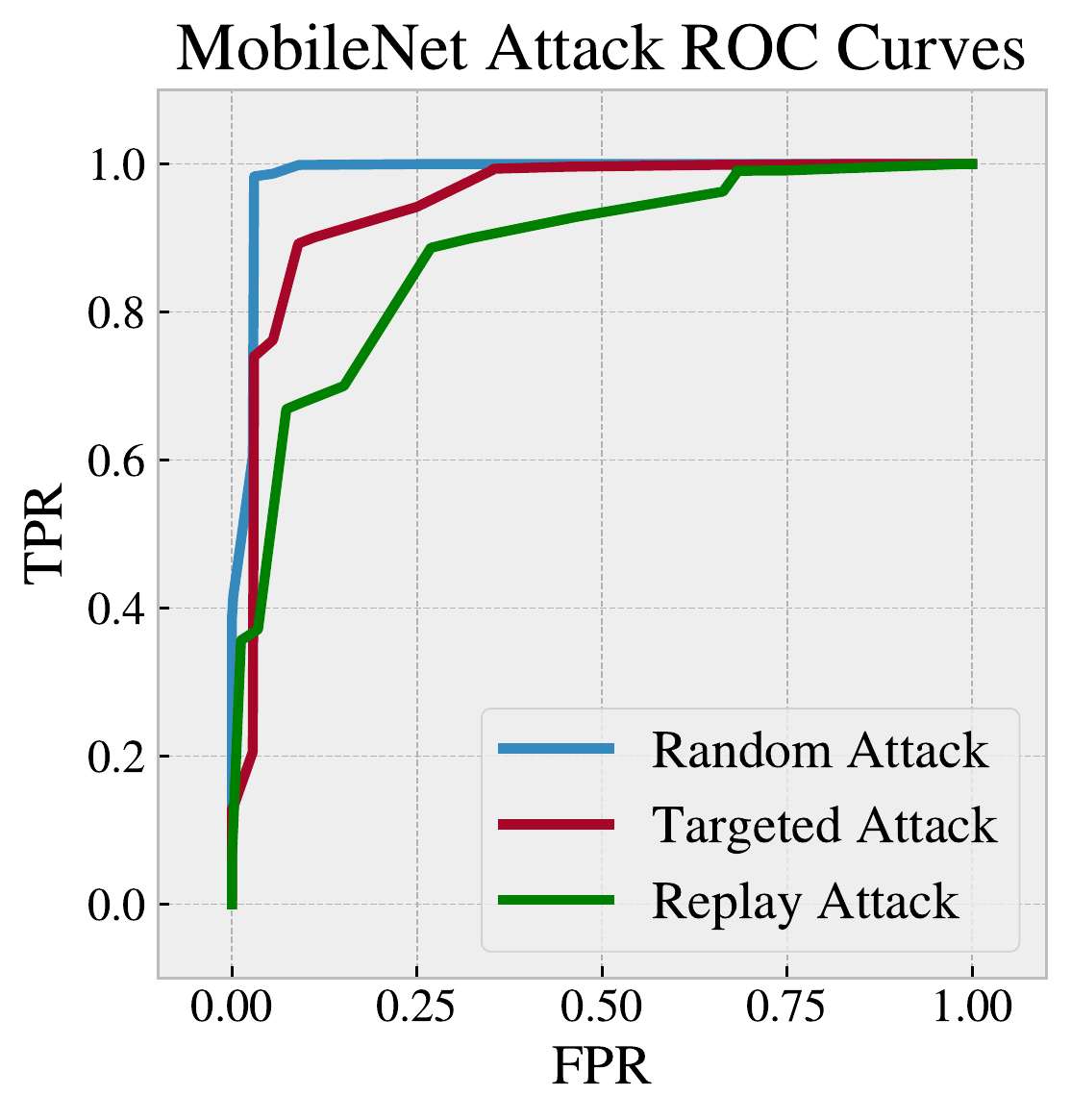}
        \caption{ROC curves for attack on MobileNet.}
        \label{subfig:rocallattackmobilenet}
    \end{minipage}
    \hspace{5mm}
    \begin{minipage}[t]{0.47\textwidth}
        \includegraphics[width=\linewidth]{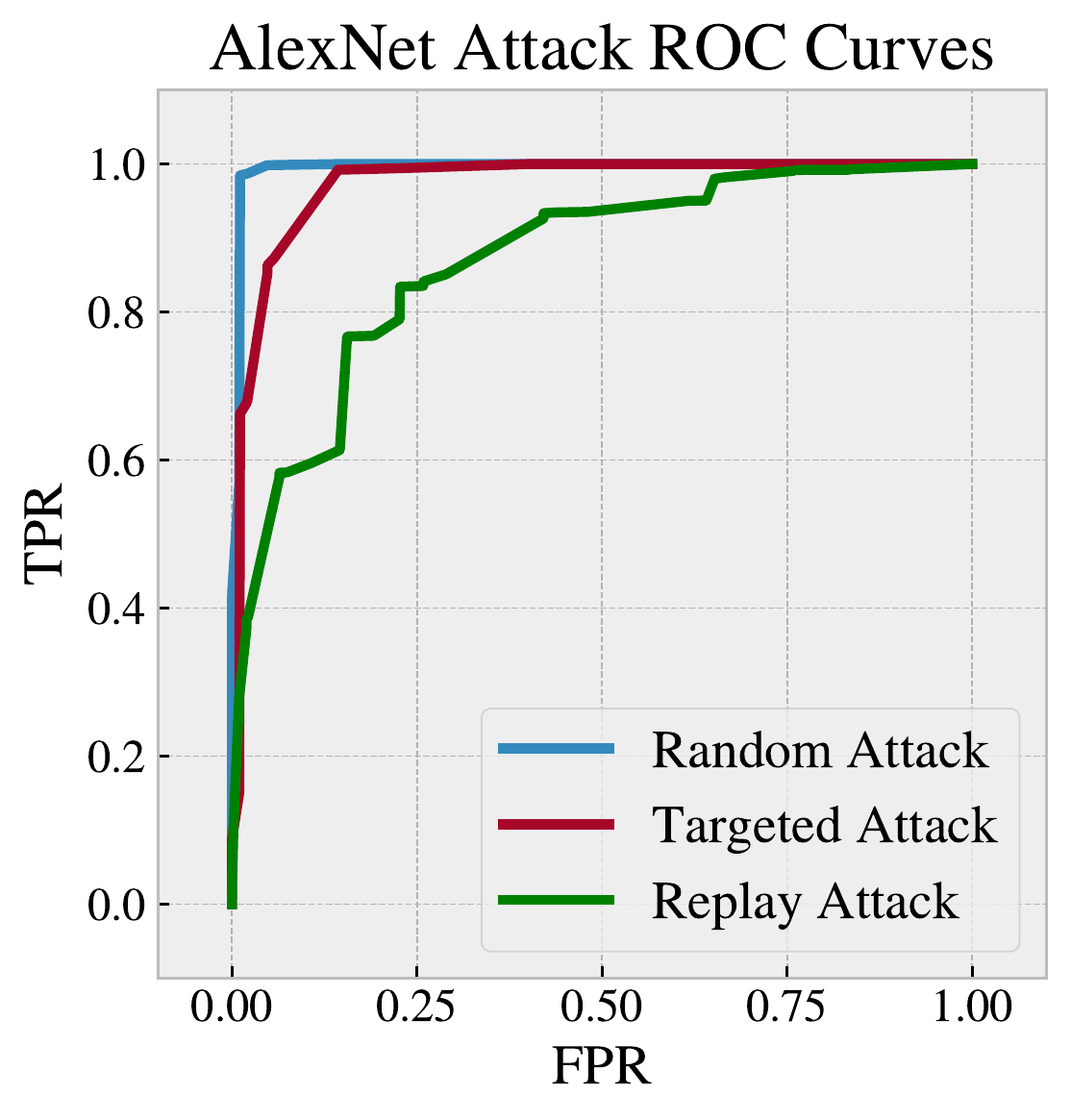}
        \caption{ROC curves for attacks on AlexNet.}
        \label{subfig:rocallattack}
    \end{minipage}
\end{figure}

The receiver operating characteristic (ROC) curves for CheckNet's detection ability for the three attack methods are shown in Figure~\ref{subfig:rocallattackmobilenet} for MobileNet and Figure~\ref{subfig:rocallattack} for AlexNet. The curves are plotted in terms of true positive rate (TPR: rate at which attacks are successfully detected) and false positive rate (FPR: rate at which legitimate inputs are incorrectly detected as attacks).
The results are shown for the following hyperparameter settings: 100 output nodes ($N_o$), 30 CrossCheck sets ($N_s$), 64-bit bithash length ($l$). The ROC curves are generated by varying the CrossCheck and HashCheck thresholds resulting in different TPR/FPR tradeoffs that make up each ROC curve. Both thresholds are set by selecting seven threshold values evenly spaced between the minimum and maximum threshold values (CrossCheck threshold $T_c$ ranges from 0 to the number of CrossCheck sets $N_s$, and HashCheck threshold $T_h$ ranges from 0 to the bithash length $l$), resulting in $49$ pairs of thresholds evaluated.  A sample is accepted only if all threshold conditions are met, and is rejected otherwise.  Threshold settings that result in a strictly worse TPR/FPR trade-off are discarded.

CheckNet provides accurate attack detection of all three attack methods, achieving AUC scores on the random, targeted, and replay attacks of 0.98, 0.95, and 0.88 with the MobileNet model, and 0.99, 0.97, and 0.87, with the AlexNet model.  The replay attack is the most difficult attack of the three to detect because the output is an otherwise valid inference output, just not for the given input sample.  As a result, the CrossCheck technique is not applicable in this case (as its job is only to verify that the output is valid, which by definition a replayed output is), so only the HashCheck technique can detect the replay attack.

\subsection{Effects of Hyperparameters on Attack Detection}
Figure~\ref{subfig:rochyperparameters} shows the effects of different hyperparameter settings for the CheckNet under a targeted attack: number of output nodes ($N_o$), number of CrossCheck sets ($N_c$), and bithash length ($l$).  
Generally, the robustness of CheckNet can be controlled by increasing 1) the number of output nodes ($N_o$) and 2) the bithash length ($l$).  Increasing the bithash length ($l$) improves attack detection (red vs blue curve), aligning with the theoretical findings in Section~\ref{sec:analysis}.  Using a smaller number of CrossCheck sets improves performance (black vs blue curve), as having a larger number of CrossCheck sets increases the number of nodes that are used in potentially conflicting ways by different CrossCheck sets (as it increases the overlap between different CrossCheck sets).  This loss in detection accuracy may be acceptable in order to gain the obfuscation benefits, as having more overlap between different CrossCheck sets makes it more difficult for an adversary to glean information regarding CrossCheck set membership.
Increasing the number of output nodes improves attack detection (black vs yellow curve), which is also also explained by the fact that there is less overlapping when more output nodes are used for the same number of CrossCheck sets.

\subsection{Robustness Against Undetected Attacks\label{sec:robustness}}
\begin{figure}[ht!]
    \centering
    \begin{minipage}[t]{0.4\textwidth}
        \includegraphics[width=\linewidth]{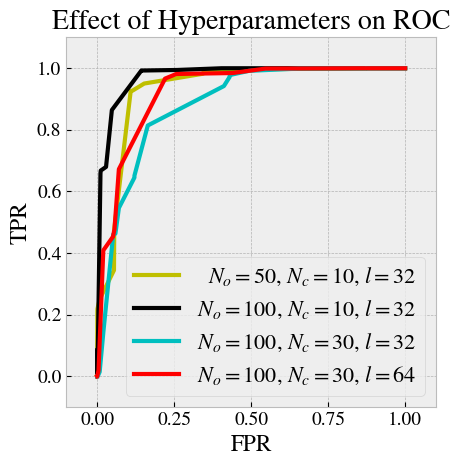}
        \caption{ROC curves under targeted attack on AlexNet for different hyperparameters: output nodes ($N_o$), CrossCheck sets ($N_c$), and bithash length ($l$).}
        \label{subfig:rochyperparameters}
    \end{minipage}
    \hspace{5mm}
    \begin{minipage}[t]{0.55\textwidth}
        \includegraphics[width=\linewidth]{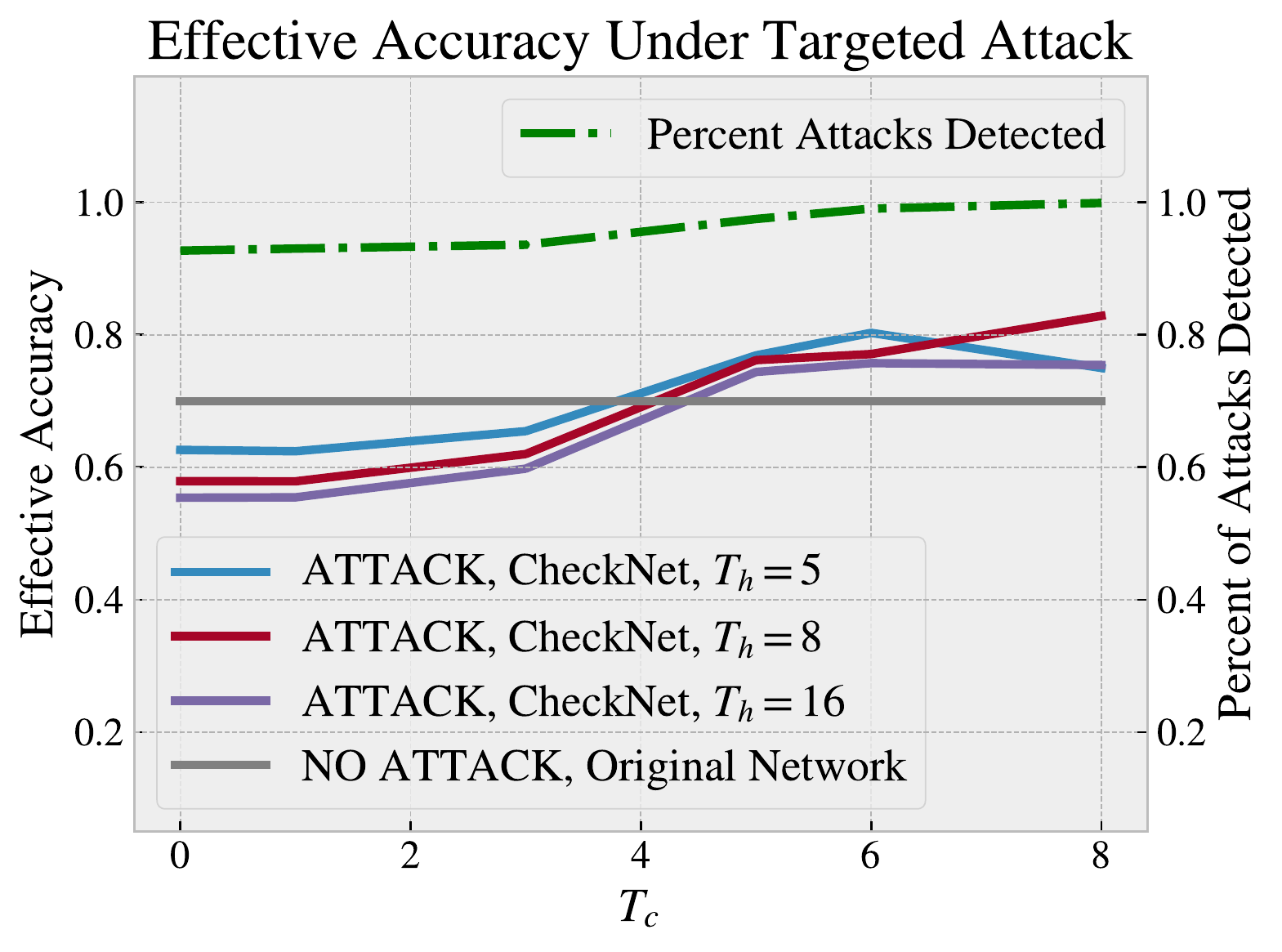}
        \caption{Effective accuracy (left y-axis) and percent of attacks detected (right y-axis) for AlexNet protected by CheckNet.  Other hyperparameters are set as follows: $N_c=10$, $N_s=100$, and $l=32$.}
        \label{fig:effectiveacc}
    \end{minipage}
\end{figure}

CheckNet users can choose a TPR/FPR tradeoff along the ROC curve that suits the application via the $T_h$ and $T_c$ hyperparameters.  Choosing a setting with a $\text{TPR}<1$ will by definition result in some attacks not being detected.  However, even in this case, CheckNet can still classify these ``undetected attack samples'' correctly in some settings due to the redundancy in the CrossCheck mechanism.  Figure~\ref{fig:effectiveacc} shows this result for different settings of $T_h$ and $T_c$ using ``effective accuracy'' (defined as the accuracy of the CheckNet protected network on the attack samples that are not rejected) as a metric under a targeted attack, compared with the accuracy of the original model with \textbf{no attack} (grey line).  For the different threshold settings, even under the targeted attack, CheckNet protected networks reject most of the attacks (dotted green line and right-y axis), and achieve similar accuracy on the attack samples it fails to reject.  Lower values of $T_c$ cause the CheckNet models to miss detecting more attacks, as shown by the green line denoting percent of attacks detected, therefore lowering the accuracy because relatively more egregious attacks that cannot be rectified via the CrossCheck majority voting are not filtered out, while higher levels filter out these harder attacks, leaving those that CrossCheck can still correctly classify even under attack.  A similar trend is also seen with regard to $T_h$, where lower, more stringent thresholds lead to higher effective accuracy than that of higher, more lenient thresholds. These results align with our theoretical analysis in Section~\ref{sec:analysis}, where we show it is harder for an attacker to launch a successful attack as $T_c$ increases or as $T_h$ decreases, resulting in a better effective accuracy overall.

\section{Analysis\label{sec:analysis}}

We derive theoretical models for robustness against attacks on the two CheckNet integrity verification mechanisms (HashCheck and CrossCheck) in order to show that it is hard to attack either of these components.  This supports the empirical results in Section~\ref{sec:robustness} showing it is hard for an attacker to avoid CheckNet's detection mechanisms and successfully alter the classification results.

\subsection{HashCheck Analysis}


We consider the attack in which the third party tries to pass the integrity check without performing the inference computation (such as the ``random'' attack described in Section~\ref{sec:results}).

\begin{definition} (Successful Attack on HashCheck)
A random attack is successful if an attacker can guess a $\boldy'$ that results in $\|\boldh_x - \boldh_{y'}\|_0 \leq T_h.$
\end{definition}

\begin{theorem}
The probability that the adversary can subvert the bithash integrity check (successful attack on HashCheck) is $1-F(l-T_h; l, \frac{1}{2})$ (where $F$ is the binomial CDF), or $1-\sum_{i=0}^{l-T_h} \binom{l}{i} \frac{1}{2}^i \frac{1}{2}^{l-i}$.
\end{theorem}
\vspace{-4mm}
\begin{proof}
Given a mapping $g$ from $\boldy'$ to $\boldh_y'$, for every bithash $\boldh_{y'}$, there exists a corresponding $\boldy'$ that maps to $\boldh_{y'}$. To guess a bithash $\boldh_{y'}$ such that $\|\boldh_x - \boldh_{y'}\|_0 \leq T_h$, an attacker needs to randomly draw $l$-bit $\boldh_{y'}$ with its corresponding $\boldy'$ such that at least $(l-T_h)$ bits match with $\boldh_x$. This is characterized by a binomial distribution with $l$ trials, each of probability $\frac{1}{2}$, i.e., the probability of a successful attack on HashCheck is equal to $1-F(l-T_h; l, \frac{1}{2})$, or $1-\sum_{i=0}^{l-T_h} \binom{l}{i} \frac{1}{2}^i \frac{1}{2}^{l-i}$.
\end{proof}

\subsection{CrossCheck}

We consider an attack to change the classification result from class $i$ to class $j$ for some $j \neq i$.  To do this, the adversary must maximize the intra-set value of the node corresponding to class $j$ in at least $T_c$ of the $N_s$ CrossCheck sets. 

\begin{definition} (Successful Attack on CrossCheck)
An attack is successful if an attacker can find $\boldy'$ that results in the majority vote is greater than $T_c.$
\end{definition}

We define the following terminology. ``Overlap'' refers to situations where nodes can be in multiple CrossCheck sets, and ``No Overlap'' refers to situations where nodes are exclusively in one CrossCheck set.  ``Knowledge'' refers to when the adversary knows which nodes are used in at least one CrossCheck set, and ``No Knowledge'' refers to when the adversary does not know if a node is used in any CrossCheck set.  In actual use, the adversary has ``No Knowledge,'' and we only consider ``Knowledge'' for the purpose of the proof. Let $A$ be an event that an attacker succeeds.

\begin{lemma}
\label{lemma:crosscheck1}
$P(A | \text{Overlap, No Knowledge}) \le P(A | \text{Overlap,  Knowledge})$
\end{lemma}
\vspace{-4mm}
\begin{proof}
An adversary selects $g$ nodes to maximize.  There are $\binom{N_s}{g}$ possible ways to successfully choose $g$ target nodes.  Without knowledge, there are $\binom{N_o}{g}$ possible sets of guesses to make.  With knowledge, there are $\binom{N_sN_c}{g}$ possible sets of guesses to make.  Because $N_sN_c < N_o$, knowledge restricts the selection space, making an attack more likely to succeed with knowledge than without.
\end{proof}
\vspace{-4mm}
\begin{lemma}
\label{lemma:crosscheck2}
$P(A|\text{Overlap, Knowledge}) \le P(A | \text{No Overlap, Knowledge})$
\end{lemma}
\vspace{-4mm}
\begin{proof}
Without overlap, selecting one of the target nodes to maximize (set above $\max(\boldy)$) has no impact on the efficacy of a second successful selection within a different CrossCheck set, as it by definition cannot be in the second set.  With overlap, in the non-zero number of cases where the first successfully selected node is is also a non-target in a second CrossCheck set, subsequently selecting the target node in this other CrossCheck set to similarly maximize will result in a tie between the two selected nodes.  This tie will have to be broken between the two nodes, which in some cases will result in the wrong node being maximized and the attack therefore being unsuccessful.  Therefore, because there are strictly more scenarios in which node selection can cause an attack to be unsuccessful in the overlap case, the overlap case has a lower probability of being successful.   
\end{proof}
\vspace{-2mm}
\begin{lemma}
\label{lemma:crosscheck3}
$P(A | \text{No Overlap, Knowledge}) = 1-F(T_c; N_s, \frac{1}{N_c})$
\end{lemma}
\vspace{-4mm}
\begin{proof}
The probability of a successful attack with this knowledge is characterized by a binomial distribution with $N_s$ (number of CrossCheck sets) trials, where each trial has success probability $\frac{1}{N_c}$, where $N_c$ is the number of nodes in each CrossCheck set. The probability that the adversary can subvert the CrossCheck integrity check with no overlap and the full knowledge of the CrossCheck sets is therefore $1-F(T_c; N_s, \frac{1}{N_c})$ (where $F(\cdot)$ is the binomial CDF), or $1-\sum_{i=0}^{T_c} \binom{N_s}{i} \frac{1}{N_c}^i \frac{N_c-1}{N_c}^{N_s-i}$.
\end{proof}
\vspace{-4mm}
\begin{theorem} 
The upper bound on the probability that the adversary can subvert the CrossCheck integrity check (Successful Attack on CrossCheck) is $1-F(T_c; N_s, \frac{1}{N_c})$ (where $F(\cdot)$ is the binomial CDF), or $1-\sum_{i=0}^{T_c} \binom{N_s}{i} \frac{1}{N_c}^i \frac{N_c-1}{N_c}^{N_s-i}$.
\end{theorem}

\vspace{-4mm}
\begin{proof}

Following Lemma~\ref{lemma:crosscheck1}, Lemma~\ref{lemma:crosscheck2}, and Lemma~\ref{lemma:crosscheck3} above, the upper bound on the probability that the adversary can successfully subvert the CrossCheck integrity check is $1-F(T_c;N_s,\frac{1}{N_c})$ (where $F_X(\cdot)$ is the binomial CDF), or $1-\sum_{i=0}^{T_c} \binom{N_s}{i} \frac{1}{N_c}^i \frac{N_c-1}{N_c}^{N_s-i}$.
\end{proof}

\section{Related Work}

CrossCheck training can be viewed as an application of dropout~\cite{srivastava2014dropout}.  Each CrossCheck set is trained by ``dropping out'' output nodes not in its set, and the CrossCheck output can be viewed as averaging a set of sub-models.  The voting method to obtain a classification from CrossCheck sets can be seen as an ensemble method, which is known to provide additional robustness and accuracy~\cite{krogh1995neural}.

Verifying inference integrity has been recently studied in limited contexts.  Recently introduced methods either frame the problem in a constrained framework (e.g., modeling the network as an arithmetic circuit) or use specialized hardware.  \cite{ghodsi2017safetynets} utilizes an interactive proof to provide proof of correctness, but is only applicable to neural networks that can be expressed as an arithmetic circuit.  This limits applicability, as it cannot use common activation functions (e.g., ReLU) except in the last layer or common pooling layers (e.g., max pooling).  In contrast, our method does not place any restrictions on the underlying model. \cite{chen2018securenets} uses a single layer at a time execution model to make modifications to the input and each layer's weight matrix in order to provide privacy and verify correctness.  It requires computing secure matrix transformations for each layer, sending the secure input and weight matrix for each layer to the third party, and returning the output for each layer to the verifier.  This adds computational and communication overhead.  In contrast, CheckNet outsources the entire inference computation to the third party, adding little computational and communication overhead.  \cite{tramer2018slalom} introduces a framework for partitioning inference calculations between a trusted execution environment (such as Intel\texttrademark~SGX~\cite{costan2016intel}) and an untrusted GPU, but requires a trusted execution environment.  In contrast, our method does not have any hardware requirements, and is therefore applicable where specialized hardware is not available (e.g., IoT devices). 

In contrast to all these methods, to our knowledge we are the first to propose a method for secure inference that is applicable to deep learning models in general without the need for specialized hardware.  As such, our proposed CheckNet method will be applicable to all devices capable of running inference computations, without adding model architecture or hardware constraints.





\section{Conclusion}

We have introduced CheckNet, a general method for verifying the integrity of inference computations performed by an untrusted third party.  CheckNet verifies that the inference computation has actually been performed and has not been manipulated by an adversary.  CheckNet will enable the expansion and scaling of inference computations to untrusted third party devices and cloud providers in a secure manner.  As machine learning is applied to increasingly critical industries such as healthcare and national safety, securing inference with methods such as CheckNet will become a critical component of all machine learning applications that rely on third parties.

The main advantages of using CheckNet for secure inference are its wide applicability, small overhead, and complete lack of prerequisites such as specialized hardware.  CheckNet is generally applicable to deep learning models (e.g., there are no theoretical limitations on model size or architecture), and requires negligible modification to the model being protected. 
CheckNet adds negligible computational overhead to the inference computation, and the verification computational overhead is very small compared to the original network size.  CheckNet also adds only negligible communication overhead, making it well suited to network-limited applications.  These overheads can be changed via easily selected hyperparameters, which also allow users flexibility in choosing an appropriate level of security (via a TPR/FPR tradeoff) based on application needs.

CheckNet provides these security benefits through two mechanisms.  HashCheck ensures that the inference output is consistent with the input.  CrossCheck ensures that the inference output has not been manipulated, doing so by obscuring the output and building in redundancy that allows the integrity of the result to be confirmed.  These mechanisms work together to thwart both ``lazy'' attacks that seek to avoid doing the computation and ``hard-working'' adversarial attacks that seek to change the outcome, both of which we have shown CheckNet can detect with high accuracy.



\medskip

\small

\bibliography{paper}
\bibliographystyle{plain}

\end{document}